\documentclass[conference]{IEEEtran}
\IEEEoverridecommandlockouts
\usepackage{cite}
\usepackage{amsmath,amssymb,amsfonts}
\usepackage{algorithmic}
\usepackage{graphicx}
\usepackage{textcomp}
\usepackage{xcolor}
\usepackage{multirow}
\usepackage{pifont}
\usepackage{bbding}
\usepackage{url}
\usepackage{hyperref}
\usepackage{booktabs}

\def\BibTeX{{\rm B\kern-.05em{\sc i\kern-.025em b}\kern-.08em
    T\kern-.1667em\lower.7ex\hbox{E}\kern-.125emX}}
\begin{document}

\makeatletter
\newcommand{\linebreakand}{%
  \end{@IEEEauthorhalign}
  \hfill\mbox{}\par
  \mbox{}\hfill\begin{@IEEEauthorhalign}
}
\makeatother

\newcommand{\todo}[1]{{\color{red}#1}}

% \title{Real-World Performance of Kolmogorov-Arnold Networks: A Benchmarking Study}
\title{A Benchmarking Study of Kolmogorov-Arnold Networks on Tabular Data}
% \title{A Benchmarking Study on Tabular Data of Kolmogorov-Arnold Networks}

\author{\IEEEauthorblockN{Eleonora Poeta}
\IEEEauthorblockA{\textit{Dip. di Automatica e Informatica } \\
\textit{Politecnico di Torino}\\
Turin, Italy \\
eleonora.poeta@polito.it}
\and
\IEEEauthorblockN{Flavio Giobergia}
\IEEEauthorblockA{\textit{Dip. di Automatica e Informatica } \\
\textit{Politecnico di Torino}\\
Turin, Italy \\
flavio.giobergia@polito.it}
\and
\IEEEauthorblockN{Eliana Pastor}
\IEEEauthorblockA{\textit{Dip. di Automatica e Informatica } \\
\textit{Politecnico di Torino}\\
Turin, Italy \\
eliana.pastor@polito.it}
\linebreakand 
\IEEEauthorblockN{Tania Cerquitelli}
\IEEEauthorblockA{\textit{Dip. di Automatica e Informatica } \\
\textit{Politecnico di Torino}\\
Turin, Italy \\
tania.cerquitelli@polito.it}
\and
\IEEEauthorblockN{Elena Baralis}
\IEEEauthorblockA{\textit{Dip. di Automatica e Informatica } \\
\textit{Politecnico di Torino}\\
Turin, Italy \\
elena.baralis@polito.it}
}

\maketitle

\begin{abstract}
Kolmogorov-Arnold Networks (KANs) have very recently been introduced into the world of machine learning, quickly capturing the attention of the entire community. However, KANs have mostly been tested for approximating complex functions or processing synthetic data, while a test on real-world tabular datasets is currently lacking. 
In this paper, we present a benchmarking study comparing KANs and Multi-Layer Perceptrons (MLPs) on tabular datasets.  The study evaluates task performance and training times. From the results obtained on the various datasets, KANs demonstrate superior or comparable accuracy and $\mathbf{F_1}$ scores, excelling particularly in datasets with numerous instances, suggesting robust handling of complex data.
We also highlight that this performance improvement of KANs comes with a higher computational cost when compared to MLPs of comparable sizes. 
\end{abstract}


\section{Introduction}
The introduction of Kolmogorov-Arnold Networks (KANs)\cite{liu2024kan} has received significant attention from the machine learning and deep learning community, quickly capturing the interest of researchers and practitioners. Inspired by the work of Andrey Kolmogorov and Vladimir Arnold~\cite{kolmogorov1956representation,arnold1957functions}, KANs represent a new class of neural network architectures. 
KANs represent a significant advancement in addressing some limitations of Multi-Layer Perceptrons (MLPs). They try to mitigate the ``black box" nature inherent in MLPs by offering an \textit{interpretable} framework. In KANs, users can inspect the network internally, understand activation functions, and interact with the network structure through pruning.
Another distinguishing feature of KANs is their fully connected structure. While this resembles MLPs, where nodes (neurons) have fixed activation functions, KANs innovate by incorporating learnable activation functions on edges (weights). 
This approach eliminates traditional linear weight matrices, replacing each weight parameter with a learnable 1D function that is typically parameterized as a spline. 
By representing functions as a sum of these simple, learnable functions, KANs significantly enhance accuracy and interpretability in function approximation tasks.

This capability makes KANs useful for scientific discovery and various complex applications, as they can accurately represent continuous functions while maintaining the interpretability of the underlying model.
Until now, KANs have mainly been evaluated on simple cases and synthetic datasets, leaving their performance unexplored on more complex, real-world data. To address this gap, we propose benchmarking KAN networks using some of the most widely utilized datasets from the UCI Machine Learning Repository\cite{Kelly:2023}.

This benchmarking study tests KANs on diverse datasets encompassing various domains and dimensionalities to assess their robustness, scalability, and practical applicability. We test KANs against their MLP counterparts.
By systematically evaluating KANs, we aim to gain a comprehensive understanding of their capabilities and limitations, ultimately contributing to their development and potential adoption in a wider range of applications.
The results indicate that KAN is a viable substitute for MLP, demonstrating competitive performance across various scenarios and excelling in more complex datasets with numerous instances, although at a higher computational cost.
% Please add the following required packages to your document preamble:

\begin{table*}
\centering
\label{tab:datasets}
\caption{Dataset characteristics. For each dataset, we report the classification type (Binary or Multiclass), the number of features, number of rows, missing values, and data types.}
\begin{tabular}{cccccc}
\toprule
\multicolumn{1}{c}{\textbf{Dataset name}} & \multicolumn{1}{c}{\textbf{Task}}  & \multicolumn{1}{c}{\textbf{Number of Features}} & \multicolumn{1}{c}{\textbf{Number Rows}}  & \multicolumn{1}{c}{\textbf{Missing values}} & \multicolumn{1}{c}{\textbf{Data Type}}\\ \midrule

\multirow{1}{*}{Breast Cancer} & Binary & 30 & 569 &  \XSolidBrush & Numerical\\ 
\multirow{1}{*}{Spam}  & Binary &47 & 4,601 & \XSolidBrush & Numerical \\
\multirow{1}{*}{Musk} & Binary & 166 & 6,598 &  \XSolidBrush & Numerical\\ 
\multirow{1}{*}{Dry Bean} & Multiclass & 16 & 13,611&  \XSolidBrush & Numerical\\
\multirow{1}{*}{Gamma Telescope} & Binary & 10 & 19,020&  \XSolidBrush & Numerical\\
\multirow{1}{*}{Adult} & Binary  & 14 & 48,842 & \Checkmark & Numerical + Categorical\\
\multirow{1}{*}{Shuttle} & Multiclass& 7 &58,000 & \XSolidBrush & Numerical\\ 
\multirow{1}{*}{Diabetes}  & Binary &21 &253,680& \Checkmark & Numerical\\ 
\multirow{1}{*}{Poker} & Multiclass & 10 & 1,025,010&  \XSolidBrush & Numerical\\

\bottomrule

\end{tabular}
\end{table*}

\section{Related work and Background}
Multi-layer Perceptrons (MLP) \cite{rumelhart1986learning, goodfellow2016deep, haykin1998neural, hornik1989multilayer} have long been a fundamental building block for constructing neural networks. Their fully connected architecture and ability to approximate complex functions, as well as their expressive power, have made them widely popular in various applications, from image recognition to natural language processing. Even if really popular, MLP architectures suffer from different drawbacks. For example, activation functions are fixed and applied to nodes (neurons). To some extent, this network rigidity may limit the model's flexibility in capturing complex relationships within the data, as it relies on predefined nonlinearities. 
Moreover, MLPs are often considered black boxes~\cite{radenovic2022neural} because their inner workings are not easily interpreted. This opacity can hinder the adoption of MLPs in fields where interpretability is critical.

\subsection{Kolmogorov-Arnold Theorem (KAT)}
The Kolmogorov-Arnold theorem (KAT) originated %in 1957 
with the foundational work of Andrey Kolmogorov~\cite{kolmogorov1956representation}, who demonstrated that any multivariate continuous function can be represented as a finite composition of several univariate continuous functions and a binary operation of addition. This result laid the theoretical foundation for the theorem. 
%In 1963, 
Vladimir Arnold~\cite{arnold1957functions} further refined Kolmogorov's theory, solidifying KAT as a robust framework for representing functions.

Specifically, a smooth multivariate continuous function on a bounded domain, $f: \left [ 0, 1 \right ]^{n} \rightarrow \mathbb{R}$, can be written as follows: 
\begin{equation} \label{eq:kan}
    f(x) = f(x_{1}, ..., x_{n}) = \sum_{q=1}^{2n+1}\Phi_{q}\left ( \sum_{p=1}^{n} \phi_{q,p}(x_{p})\right )
\end{equation}
with $\phi_{q,p} : \left [ 0, 1 \right ] \rightarrow \mathbb{R}$ and $\Phi_{q} : \mathbb{R}\rightarrow \mathbb{R}$ being the univariate functions.

Despite its theoretical significance, KAT was not initially seen as directly applicable to neural networks. In \cite{girosi1989Irrelevant}, the authors address the challenge of representing nonlinear mappings with simpler functions of fewer variables in neural networks, reviewing the Kolmogorov-Arnold theorem. They conclude that KAT is irrelevant for neural networks because these networks can approximate multivariable functions using one-input, one-output units without relying on the specific representation properties outlined by Kolmogorov's theorem. 

Additionally, the theorem is based on using only two hidden layers and a limited set of activation functions, which is impractical for modern neural networks. 
Contemporary neural network architectures can involve hundreds of hidden layers and thousands of neurons within those layers. As a result, deeper networks with nonlinear activation functions tend to perform significantly better for function approximation tasks.

% Please add the following required packages to your document preamble:

\begin{table*}[ht]
\centering
\label{tab:performances}
\caption{Perfomance results of KAN and MLP across the datasets.
Each metric is expressed as average$\pm$standard deviation.
The selected accuracy, F1-score, precision, and recall are the highest among the average metrics from five runs across all parameter configurations. The FPR and FNR are calculated for binary classification datasets only. The training time in seconds refers to the average expired time in training for each configuration.
}
\begin{tabular}{ccccccccc}

\toprule
\multicolumn{1}{c}{\textbf{Dataset}}   & \multicolumn{1}{c}{\textbf{Model}} & \multicolumn{1}{c}{\textbf{Accuracy}} &\textbf{ F1 score} & \multicolumn{1}{c}{\textbf{Precision}} & \multicolumn{1}{c}{\textbf{Recall}} & \multicolumn{1}{c}{\textbf{FPR}} & \multicolumn{1}{c}{\textbf{FNR}}  & \multicolumn{1}{c}{\textbf{Training time (s)}} \\ \midrule

\multirow{2}{*}{Breast Cancer} & KAN   & 94.56$\pm$1.569& 78.69$\pm$1.692 & 95.03$\pm$1.791 & 93.43$\pm$1.685 & 1.97$\pm$1.606& 11.163$\pm$2.548 & 0.09\\
& MLP   & \textbf{96.84$\pm$0.480} & \textbf{80.44$\pm$0.423} & \textbf{96.64$\pm$0.540} & \textbf{96.64$\pm$0.536}& 2.53$\pm$0.630& 4.18$\pm$1.040 & 0.06\\
\midrule
\multirow{2}{*}{Spam} & KAN   & \textbf{94.09$\pm$0.238 }& 78.31$\pm$0.245 & \textbf{94.10$\pm$0.249} & \textbf{93.78$\pm$0.247} & 4.18$\pm$0.309 & 8.26$\pm$0.421 & 0.69\\
& MLP   & 93.96$\pm$0.182 &\textbf{ 79.19$\pm$0.18}8 & 93.93$\pm$0.180 & 93.69$\pm$0.195 &4.52$\pm$0.133 & 8.11$\pm$0.292 & 0.38\\
\midrule
\multirow{2}{*}{Musk} & KAN   & \textbf{92.45$\pm$0.358}  & \textbf{83.62$\pm$0.992} & 93.71$\pm$1.492 & 89.86$\pm$1.943& 1.53$\pm$0.397& 18.76$\pm$3.571 & 0.97\\
& MLP   & 90.44$\pm$1.125 & 76.47$\pm$4.119& \textbf{97.05$\pm$0.577} & \textbf{91.07$\pm$1.223}& 0.45$\pm$0.134& 17.42$\pm$2.383 & 0.43\\
%& DT   & 94,97$\pm$0.115 & 93.78$\pm$0.107 \\
%& RF   & 97.89$\pm$0.149 & 95.83$\pm$0.311\\
%& RF\_60   & 97.85$\pm$0.231 & \textbf{97.79$\pm$0.246} \\ 
\midrule
\multirow{2}{*}{Dry Bean} & KAN   & \textbf{92.80$\pm$0.158} & \textbf{92.82$\pm$0.162}& \textbf{92.87$\pm$0.169} & \textbf{92.80$\pm$0.158} & - & - & 1.92\\
& MLP   & 92.75$\pm$0.088 & 92.77$\pm$0.085 & 92.81$\pm$0.076 & 92.75$\pm$0.088 & - & - & 0.86\\
%& DT   & 87.58$\pm$0.000& 89.59$\pm$1.581 \\
%& RF   & 92.35$\pm$0.539 & 92.36$\pm$0.147\\
%& RF\_15   & \textbf{92.46$\pm$0.172} & 92.25$\pm$0.173 \\ 
\midrule
\multirow{2}{*}{Gamma Telescope} & KAN   & \textbf{86.94$\pm$0.313 } & \textbf{71.01$\pm$0.331} & \textbf{87.03$\pm$0.470} & \textbf{83.97$\pm$0.287} & 5.91$\pm$0.487& 26.15$\pm$0.463 & 2.74\\
& MLP   & 85.94 $\pm$0.042 & 69.56$\pm$0.049 & 85.83$\pm$0.083 & 82.12$\pm$0.061& 6.05$\pm$0.127 & 29.71$\pm$0.213 & 1.45\\
%& DT   & 82.28$\pm$0.000 & 83.71$\pm$0.000 \\
%& RF   & 87.87$\pm$0.111 & 86.32$\pm$0.126\\
%& RF\_10   & 86.91$\pm$0.358 & 85.07$\pm$0.404 \\ 
\midrule
\multirow{2}{*}{Adult} & KAN   & \textbf{85.93 $\pm$0.061}  & \textbf{80.24$\pm$0.231}& \textbf{82.71$\pm$0.249 }& 78.52$\pm$0.443& 6.41$\pm$0.254& 37.39$\pm$1.180 & 6.66\\
& MLP   & 85.72$\pm$0.119  &79.87$\pm$0.313 & 82.47$\pm$0.070 & \textbf{78.89$\pm$1.995}& 6.40$\pm$0.419& 36.57$\pm$1.295 & 3.47\\
%& DT   & 84.37$\pm$0.007 & 76.24$\pm$0.013 \\
%& RF   & 85.53$\pm$0.105&  79.26$\pm$1.445\\ 
%& RF\_24   & 85.16$\pm$0.158& 79.18$\pm$0.263 \\
\midrule
\multirow{2}{*}{Shuttle} & KAN   & \textbf{99.76$\pm$0.013}  & \textbf{99.70$\pm$0.021} & \textbf{99.72$\pm$0.056} & \textbf{99.76$\pm$0.013 } & - & - & 8.39\\
& MLP   & 99.62$\pm$0.011  & 99.47$\pm$0.016 & 99.49$\pm$0.023 & 99.62$\pm$0.011 & - & - & 4.22\\
%& DT   & \textbf{99.79$\pm$0.004 }&  \\
%& RF   & 99.55$\pm$0.162 &  \\ 
%& RF\_3   & 99.84$\pm$0.022 & $\pm$ \\
\midrule
\multirow{2}{*}{Diabetes} & KAN   & \textbf{86.80 $\pm$0.028}  & \textbf{58.38$\pm$0.728} & 73.00$\pm$0.228 &\textbf{ 56.69$\pm$0.503} &  1.74$\pm$0.166& 84.86$\pm$1.170 & 38.74\\
& MLP   & 86.73$\pm$0.194  & 55.92$\pm$3.860 & \textbf{74.06$\pm$1.070} & 55.19$\pm$2.198& 1.26$\pm$0.579 & 88.35$\pm$4.972 & 20.34\\
%& DT   & 86.61$\pm$0.000 & 55.38$\pm$0.894 \\
%& RF   & 85.41$\pm$0.017& 47.89$\pm$1.901 \\ 
%& RF\_30   & 85.25$\pm$0.028& 59.92$\pm$0.048\\
\midrule

\multirow{2}{*}{Poker} & KAN   & \textbf{99.91$\pm$0.038}  & \textbf{99.91$\pm$0.039} &  \textbf{99.91$\pm$0.039}& \textbf{99.91$\pm$0.038}&- & - & 34.76\\
& MLP   & 92.44$\pm$0.209  & 89.07$\pm$0.610 & 86.24$\pm$1.186 & 92.44$\pm$0.209 & - & - & 13.98\\ 
%& DT   & 55.44$\pm$0.000 & 51.75$\pm$0.000 \\
%& RF   & & $\pm$ \\ 
%& RF\_3   & 60.88$\pm$0.929& $\pm$ \\

\bottomrule

\end{tabular}
\end{table*}

\subsection{Kolmogorov-Arnold Networks (KAN)}
Liu et al.~\cite{liu2024kan} are the first to propose Kolmogorov-Arnold Networks (KANs) as promising alternatives to MLPs, addressing the previously mentioned limitations. 
Their work differentiates from previous research by identifying the similarity between MLPs and networks using the Kolmogorov-Arnold Theorem, which they term Kolmogorov-Arnold Networks (KANs). 
Typically, an MLP is characterized by its multiple stacked layers. 
Therefore, the focus is on identifying what constitutes a ``layer" within KAN architectures. 
From equation \ref{eq:kan}, the network is structured as a two-layer neural network with activation functions on the edges rather than the nodes (where a simple summation is performed) and with a width of \(2n + 1\) in the middle layer. 
This clarifies the concept of deeper Kolmogorov-Arnold representations by simply stacking more KAN layers.
One distinctive feature of KAN networks is the absence of traditional weights of neural networks. 
In KANs, every ``weight'' is effectively represented as a small function. 
Unlike traditional neural networks where nodes apply fixed non-linear activation functions, each edge in a KAN is characterized by a learnable activation function.
%that receives inputs and produces outputs dynamically.
This architectural paradigm allows KANs to operate more flexibly and adaptively than conventional approaches, potentially allowing the modeling of complex relationships in data.

\section{Benchmarking Evaluation}
% piccola intro su cosa troviamo qui dentro
This section outlines the methodology employed in this study for benchmarking. We provide an overview of the benchmarking process, including a description of the datasets used, the testing procedures, the metrics examined, and the configurations of the KAN and MLP architectures evaluated.

\subsection{Benchmarking methodology}
The study benchmarks KANs by comparing their performance and effectiveness with those of traditional MLPs. 
We aim to evaluate whether KANs could be a viable alternative and motivate their adoption. 
For the analysis, we use real-world datasets to explore the practical applications and effectiveness of KANs in scenarios commonly encountered in everyday data analysis tasks.

The benchmarking process evaluates the two models across ten architectural configurations, ensuring 
each configuration has a comparable number of parameters for both models.
This balanced configuration allows us to determine which model demonstrates superior performance under equivalent conditions.
We progressively increase the number of parameters and evaluate how performance varies.

Regarding data handling, each dataset undergoes uniform preprocessing tailored to its specific data type, ensuring consistency and reliability in the evaluation process.

\begin{figure*}
    \centering
    \includegraphics[width=\linewidth]{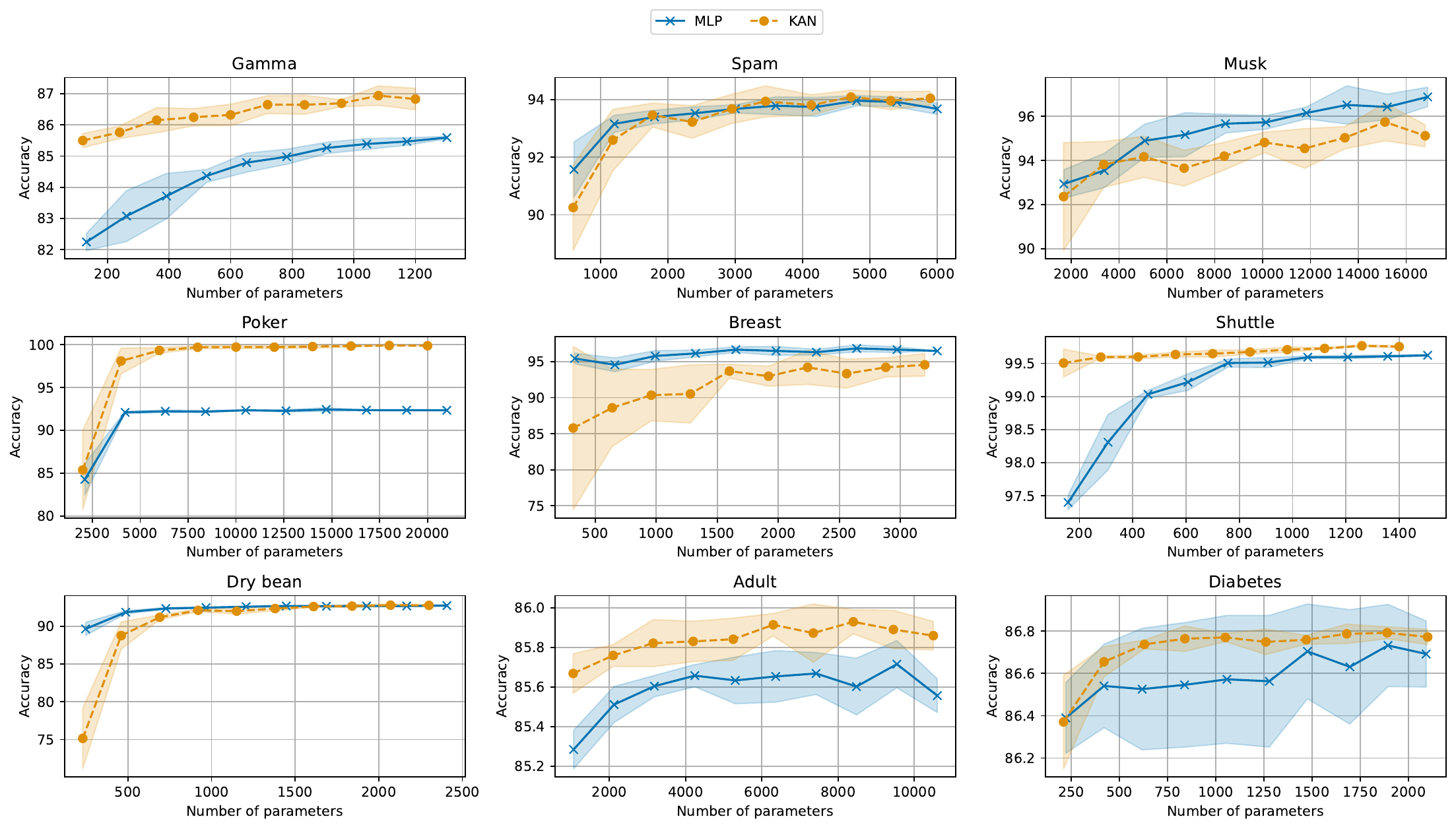}
    \caption{Accuracy scores of KANs and MLPs as the number of parameters increases. For each dataset, we report the average accuracy across five runs for each parameter count.}
    \label{fig:acc_results}
\end{figure*}

\subsection{Datasets}
\label{sec:dataset}
The benchmarking conducted in this study utilizes datasets from the UCI Machine Learning Repository\cite{Kelly:2023}. 
This selection was made to evaluate the performance of KANs on real-world datasets, which are widely used within the data science community.
We select the datasets considering different aspects and scenarios, such as testing KAN in the \textit{many instances} scenario or the \textit{few features} scenario. 
The considered datasets are as follows:

\begin{itemize}
    \item \textbf{Breast Cancer Wisconsin Diagnostic (Breast Cancer)}:\cite{misc_breast_cancer_wisconsin_(diagnostic)_17} This dataset comprises features computed from digitized images of fine needle aspirates (FNA) of breast masses. It is generally used to predict whether a breast mass is malignant or benign.
     \item \textbf{Spambase (Spam)}:\cite{misc_spambase_94} This dataset is utilized for email spam detection, containing a collection of email messages categorized as spam or non-spam. It aids in the development of models for automatic spam filtering systems. %, facilitating the task of spam classification.
     \item \textbf{Musk}\cite{misc_musk_(version_2)_75}: This dataset contains descriptions of molecules to distinguish between musks (active) and non-musks (inactive). It is used primarily for classification tasks in chemoinformatics and bioinformatics. We use the $2^{nd}$ version of this dataset as it contains more instances.
     \item \textbf{Dry Bean} \cite{misc_dry_bean_602}: It includes morphological features of seven types of dry beans. The dataset is used to classify the types of dry beans.
    \item \textbf{MAGIC Gamma Telescope (Gamma Telescope)}: \cite{misc_magic_gamma_telescope_159} This dataset consists of measurements from a gamma-ray telescope. It is utilized to classify gamma-ray sources based on their spectral characteristics, aiding in astrophysical research.
    \item \textbf{Adult}:\cite{misc_adult_2} Also known as the  ``Census Income" dataset, it contains information extracted from the 1994 U.S. Census database. It is used for predicting whether a person makes over \$50,000 a year.
    \item \textbf{Statlog (Shuttle)}:\cite{misc_statlog_(shuttle)_148} This dataset is used for space shuttle component identification and status prediction. It includes various sensor measurements for classifying the status of the shuttle's components.
    \item \textbf{CDC Diabetes Health Indicators (Diabetes)}: This dataset contains behavioral risk factors for diabetes derived from the Behavioral Risk Factor Surveillance System (BRFSS) 2015 survey data by the CDC. It is used for predicting diabetes status.
    \item \textbf{Poker Hand (Poker)}: \cite{misc_poker_hand_158} This dataset is used to determine the poker hand held by a player based on five cards drawn from a standard deck. It is used to classify the type of poker hand (e.g., one pair, two pairs, three of a kind, etc.).

\end{itemize}

These datasets present various challenges, such as varying numbers of instances and features, the presence of missing values, and a range of classification tasks. 
This diversity allows for the thorough evaluation of KANs' performance across multiple real-world scenarios. 
We summarize the main characteristics of the datasets in Table~\ref{tab:datasets}.

\subsection{Architectures configuration}

For the experiments, we employed an efficient implementation of the KAN architecture detailed in~\cite{Blealtan2024}. This version addresses some performance issues of the implementation proposed by~\cite{liu2024kan}.
% , specifically, the challenges arising from expanding all intermediate variables to accommodate various activation functions. 
In our experiments, we use a basic KAN model with a single input layer, an intermediate layer configured with ten distinct dimensions of $k$ nodes where $k \in \left [ 1, 10 \right ]$, and an output layer for classification.

Similarly, the MLP model used in our study includes an intermediate layer of $k$ nodes, with $k \in \left [ 10, 100 \right ]$, ensuring comparability in the number of parameters with the KAN model. As an activation function, we use the Sigmoid Linear Unit (SiLU)\cite{elfwing2017sigmoidweighted, ramachandran2017searching}. 

It should be noted that we expanded the range of nodes $k$ for the Poker dataset due to the unsatisfactory results we encountered with the MLP. The new number of $k$ nodes for both architectures is increased by a factor of 10 compared to the previously stated values.

In each case, we ensure a fair and meaningful comparison by evaluating architectures with the same number of parameters under equivalent experimental conditions.

 \begin{figure*}
    \centering
    \includegraphics[width=\linewidth]{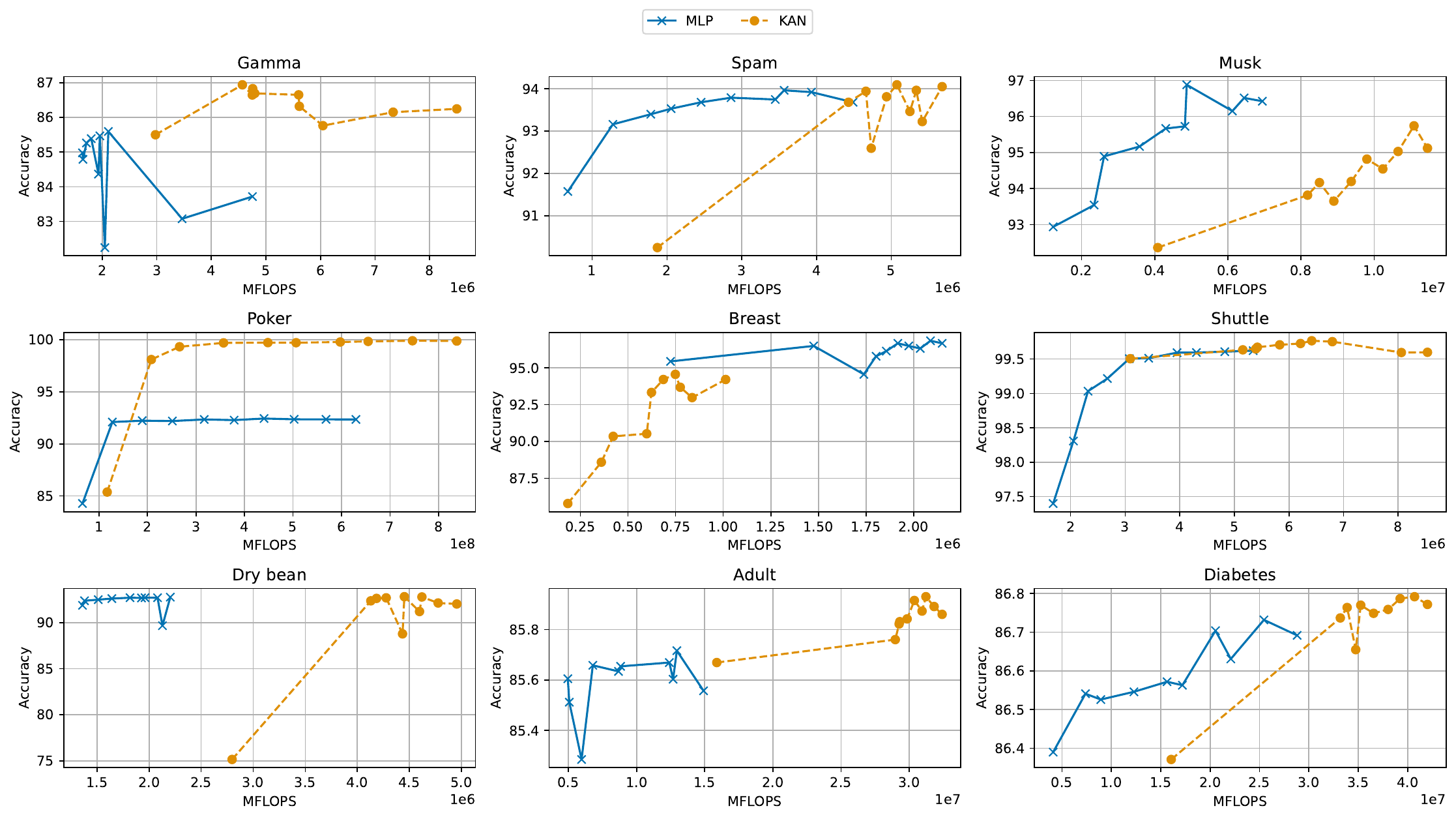}
    \caption{Number of FLOPS. The number of FLOPS performed by each model as the number of parameters increases is reported in megaflops (MFLOPS).}
    \label{fig:flop}
\end{figure*}

\subsection{Metrics}
\label{ssec:metrics}
Our assessment employs several key task performance metrics: accuracy score, $F_1$ score, precision, recall, False Positive Rate (FPR), and False Negative Rate (FNR). For datasets with a multiclass target, the $F_1$ score, precision and recall are computed using a weighted average \footnote{We note that the definition of the weighted recall coincides with that of the accuracy}. Instead, for datasets with binary targets, we report the macro (i.e., unweighted) average. 
%These metrics have been chosen to provide a thorough evaluation of the models' capabilities across different dimensions of classification performance and efficiency.

To evaluate the time efficiency of each model, we measure the training duration for every configuration used in our experiments.
At test time, to measure the number of FLOPS, we use the Python implementation \cite{python-papi} of the PAPI \cite{papi} library. 
 
\section{Experiments}
In this section, we present the experiments conducted on KAN and MLP to empirically evaluate their effectiveness across diverse tabular datasets. It is well-known in the literature that tree-based methods such as Random Forests \cite{breiman2001random} or gradient-boosted trees \cite{Chen_2016} typically achieve better performance on tabular data\cite{grinsztajn2022tree}. However, because the KAN model was introduced specifically as an alternative to MLP, we have chosen to focus our analysis on comparing KAN with MLP and, therefore, have not included tree-based methods in our evaluation.

We assessed task performance (as detailed in Subsection \ref{ssec:metrics}), training time per epoch and number of floating-point operations (FLOPS) required for task completion. We ran each experiment 5 times with different seeds. In each run, every model is trained for 10 epochs using AdamW \cite{loshchilov2019decoupled} optimizer with a learning rate of $10^{-2}$. Additionally, we apply an exponential decay to the learning rate, with a decay factor $0.8$.
The code to reproduce our experiments is available at \url{https://github.com/eleonorapoeta/benchmarking-KAN}

\subsection{Results}
This section presents the evaluation results of the KAN and MLP models using the datasets listed in Table \ref{tab:datasets}. We report the performance results and training time in Table \ref{tab:performances}.

%We report FPR and FNR values for the model that achieves the highest overall accuracy on each binary dataset.
The values in the table for $F_1$ score, precision, recall, FPR, and FNR correspond to the model with the highest overall accuracy for each dataset.

Across the range of datasets examined in this study, the KAN consistently achieves better performance w.r.t. MLP. When evaluating metrics such as accuracy and $F_1$ score, KAN's results consistently outperform those achieved by MLPs, showing an improvement by several percentage points. Indeed, KAN's performances are particularly good in datasets with larger volumes of instances, such as the Poker dataset. This trend underscores KAN's effectiveness in handling various data complexities and task demands.

%The comparative analysis reveals KAN's remarkable advantage, particularly in datasets with larger volumes of instances, such as the Poker dataset.Here,  KAN surpasses MLPs by approximately 30\% in accuracy and 40\% in $F_1$ score, underscoring its superior efficacy in managing large-scale datasets. This advantage is also evident in the Breast cancer dataset, which has the fewest instances and features among those tested. Although the performance difference is less pronounced here, KAN still performs less than MLP.

Regarding precision and recall, the KAN model generally performs better, although no consistent trend is evident.

%Notably, the results for FPR  and FNR are of interest. As shown in the table, both the MLP and KAN models maintain a consistently low FPR, indicating high accuracy in identifying negative instances in various datasets. In contrast, the FNR exhibits greater variability. For example, in the Diabetes dataset, the FNR is significantly high, suggesting that the models often fail to correctly identify positive instances. This is a critical observation, as a high FNR in medical diagnostics means that many true cases go undetected, which could lead to serious consequences.

Figure \ref{fig:acc_results} shows the accuracy of KAN and MLP models as the number of parameters increases. Across most datasets, both models generally improve in accuracy as the number of parameters increases. The degree of improvement and the performance gap between models can vary significantly depending on the dataset. For example, the KAN model shows substantial advantages in datasets like Poker and Musk, while differences are minimal in datasets like Dry Bean and Shuttle. %Overall, the improvement is more pronounced for KAN models than the MLP ones.

Another metric we use to evaluate the effectiveness of KANs is the number of FLOPS, expressed in megaflops (MFLOPS). Figure \ref{fig:flop} shows, for each dataset, the performance of the models, as the number of operations used increases. The results show that in most cases KANs perform a larger number of operations w.r.t. MLPs, despite using comparable numbers of parameters. This indicates that KANs achieve higher performance at the cost of a higher number of operations, for a fixed budget of parameters. This fact can be explained thanks to the richer activation functions used in KANs, which allow for a better, although more computationally expensive modeling of the problem.

% the curve representing the number of FLOPS for KAN is shifted to the right relative to the MLP curve, indicating that KAN has a higher number of FLOPS. These instances correspond to the cases where KAN's accuracy surpasses MLP's. Therefore, it can be inferred that KAN's higher accuracy is accompanied by a significantly higher number of FLOPS than MLP, even with the same number of parameters. 
%This suggests that KAN's superior accuracy is likely due to performing substantially more operations, even with the same parameters.}

%the trend in the number of FLOPS generally parallels that of accuracy. More FLOPS can mean more computational power, which might lead to better performance, but it also depends on how efficiently the model uses its computations. 

\section{Conclusion}
In this paper, we present a benchmarking study of the recently introduced Kolmogorov-Arnold Networks (KANs). Specifically, we measure the task performance and time efficiency of KANs compared to MLPs using tabular datasets, extending the analysis beyond the synthetic data used in the original paper.

From the results, we can conclude that KANs are a promising alternative to MLPs, demonstrating comparable performance across various tasks. Notably, KANs excel in datasets with many instances, indicating their potential to handle complex datasets effectively. We highlighted how the improvements in the performance of KANs come at a higher computational cost, as expected, given the adoption of more complex activation functions. 
%will focus on delving into the interpretability of KANs, building on their great performance. Moreover, 
Future research could focus on extending the currently proposed investigation to encompass regression tasks and diverse data types to improve the understanding of how KANs can be effectively applied in real-world contexts.

%\section*{Acknowledgment}

\bibliographystyle{IEEEtran}
\bibliography{IEEEabrv,bibliography}

\end{document}